\documentclass[10pt,twocolumn,letterpaper]{article}

\usepackage{wacv}
\usepackage{times}
\usepackage{epsfig}
\usepackage{graphicx}
\usepackage{amsmath}
\usepackage{amssymb}

\usepackage{cite}   

\usepackage{enumitem}
\usepackage{graphicx}
\usepackage{subfloat}
\usepackage{tabularx}
\usepackage{adjustbox}

\usepackage{multirow}
\usepackage[flushleft]{threeparttable}
\usepackage{subfigure}
\usepackage{color, colortbl}

\usepackage{hyperref}
\hypersetup{
	colorlinks   = true,
	citecolor    = green
}
\hypersetup{linkcolor=red}

\definecolor{LightCyan}{rgb}{0.88,1,1}
\definecolor{Gray}{gray}{0.95}

\newcommand{\figref}[1]{Fig.~\ref{#1}}
\newcommand{\tabref}[1]{Table.~\ref{#1}}
\newcommand{\eref}[1]{Eq.~(\ref{#1})}
\newcommand{\sref}[1]{Sec.~\ref{#1}}

\wacvfinalcopy 

\ifwacvfinal\fi
\begin{document}

\title{StairNet: Top-Down Semantic Aggregation for Accurate One Shot Detection}

\author{Sanghyun Woo \\
KAIST\\
{\tt\small shwoo93@kaist.ac.kr}
\and
Soonmin Hwang \\
KAIST\\
{\tt\small jjang9hsm@kaist.ac.kr}
\and
In So Kweon \\
KAIST\\
{\tt\small iskweon@kaist.ac.kr}
}

\maketitle
\ifwacvfinal\thispagestyle{empty}\fi

\begin{abstract}
One-stage object detectors such as SSD or YOLO already have shown promising accuracy with small memory footprint and fast speed. However, it is widely recognized that one-stage detectors have difficulty in detecting small objects while they are competitive with two-stage methods on large objects. In this paper, we investigate how to alleviate this problem starting from the SSD framework. Due to their pyramidal design, the lower layer that is responsible for small objects lacks strong semantics(e.g contextual information). We address this problem by introducing a feature combining module that spreads out the strong semantics in a top-down manner. Our final model StairNet detector unifies the multi-scale representations and semantic distribution effectively. Experiments on PASCAL VOC 2007 and PASCAL VOC 2012 datasets demonstrate that StairNet significantly improves the weakness of SSD and outperforms the other state-of-the-art one-stage detectors.
\end{abstract}


\section{Introduction}

Convolutional Neural Networks(CNNs) have significantly pushed the performance of visual recognition tasks such as image classification\cite{krizhevsky2012imagenet, simonyan2014very, szegedy2015going, he2016deep, huang2017densely}, semantic segmentation\cite{chen15deeplab, long2015fully, noh2015learning, yu2016multi} and object detection\cite{girshick2014rich, girshick2015fast, ren2015faster, liu2016ssd, redmon2016you}. Thanks to the representation power of the CNNs, features learned by neural networks in an end-to-end, data-driven manner have provided dramatically better results than hand-crafted features. Hence, it is not surprising that most of the recent researches on visual recognition are based on ‘network engineering’ rather than ‘feature engineering’\cite{xie2017aggregated}. Designing the better network architectures became a critical issues on a broad array of vision problems and among them object detection is one of the fastest-moving areas.

Recent object detectors that are based on the CNN can be divided into two streams. The first is two-stage detectors popularized by R-CNN\cite{girshick2014rich}, where sparse regions were proposed in the first stage then followed by a second stage for refinement. They guarantee high accuracy but is not suitable for real-time applications due to the high memory usage and slow speed, e.g 5FPS. On the other hand, the one-stage approaches such as the Single Shot Detector(SSD)\cite{liu2016ssd} or You Only Look Once(YOLO)\cite{redmon2016you, redmon2017yolo9000} directly predict the output without region proposal module. They are fast and simple to train end-to-end. However, it is shown that they produce a low quality bounding boxes, hence, results in a failure localization of small objects or occluded objects.\cite{ren2017accurate} We focus to alleviate this issues by carefully dissolving recent ideas into the network design.

\begin{figure}[t]
\def\arraystretch{0.5}
\begin{tabular}{@{}c@{\hskip 0.01\linewidth}c@{\hskip 0.01\linewidth}c@{\hskip 0.01\linewidth}}

\includegraphics[width=0.5\linewidth]{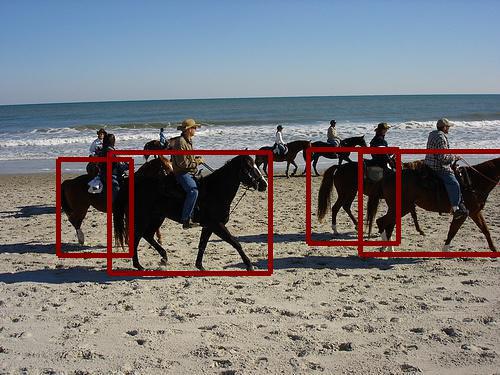}&
\includegraphics[width=0.5\linewidth]{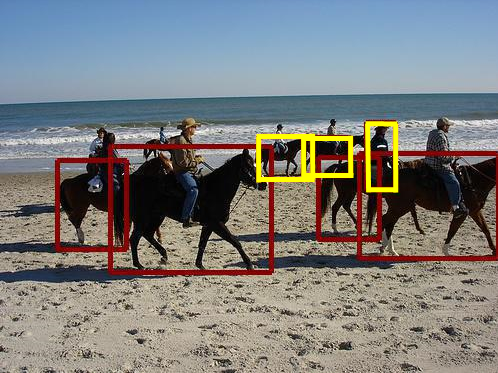}&\\
\includegraphics[width=0.5\linewidth]{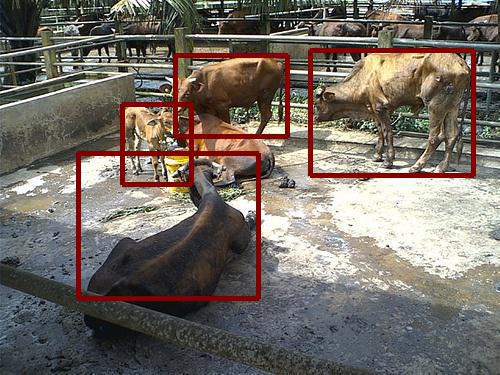}&
\includegraphics[width=0.5\linewidth]{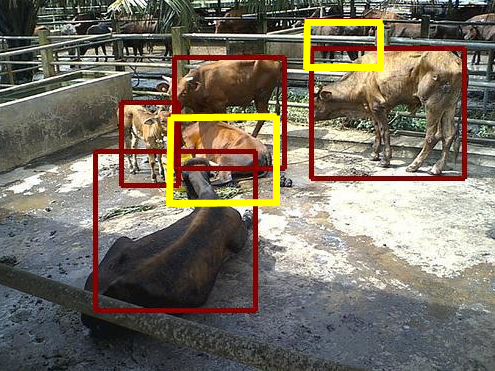}&\\
{\small (a)} & {\small (b) }\\
\end{tabular}
\caption{ {\bf The Detection Output of original SSD and StairNet.} Given the input image, (a) shows the output of SSD  and (b) shows the output of StairNet. As shown in the image, one-stage detectors\cite{liu2016ssd, redmon2016you} have poor performance on objects that require significant context information(e.g. small, overlapped and truncated).}
\label{fig:teaser}
\end{figure}

Our first design principle is motivated by \cite{felzenszwalb2010object, liu2016ssd}. Detecting objects of various scales has long been a demanding challenge. Prior to the advent of the CNNs, image pyramids were proposed as a solution. For example, deformable part models(DPM)\cite{felzenszwalb2010object} use a multi-scale images to produce a multi-scale features then the filters slide densely on top of the feature pyramid. Recent top-ranked detectors in the benchmarks\cite{everingham2010pascal,lin2014microsoft} also use multi-scale images for the training and testing. Despite the promising results of image pyramids, computation time increases considerably and memory usage gets high. Instead, we draw on a recent approach \cite{liu2016ssd}, especially adopting the SSD-style pyramid. The multi-scale feature maps are already constructed by a subsampling layers in a deep CNN. SSD have shown the effectiveness of this cost-free inherent representations for object detection. However, the original SSD-style pyramid misses to exploit semantically strong information which is critical for small object detection. While following the SSD, we augmented the pyramid with our feature combining module to boost performance.

It has been widely recognized that the contextual information is decisive on detecting visually impoverished objects (e.g. small, truncated and occluded objects). Many of recent detectors are proposed to use contextual information.\cite{bell2016inside, fu2017dssd, lin2017feature, ren2017accurate}. However due to their complex\cite{bell2016inside, ren2017accurate} and heavy architecture{\cite{fu2017dssd}}, the network shows slow inference time and is not able to be trained in end-to-end\cite{fu2017dssd}. To address these issues, we propose a simple module that propagates semantically strong features, which contain contextual information, from top to bottom in the network. In this respect, FPN\cite{lin2017feature} is similar to ours in that the final goal is to construct a multi-scale feature maps with small semantic gaps. \cite{lin2017feature} combines high level features with low level features, enabled by a nearest neighbor upsampling and lateral connections. Instead, we adopt different components and carefully designed the top-down feature combining module that significantly improves the weakness of SSD. Our final model is called StairNet that     unifies multi-scale representations and semantic distribution in an efficient way. StairNet is an end-to-end, one-stage detector which outperforms current state-of-the-art one-stage detectors.
 
 Our main contributions are summarized as follows:
 
\begin{itemize}
\item We propose StairNet framework that effectively unifies multi-scale representations and semantic distribution.
\item We conduct extensive ablation experiments and introduce a set of effective design choices for feature combining.
\item We show that StairNet can acheive state-of-the-art performance on two standard benchmarks (PASCAL VOC 2007 and PASCAL VOC 2012) without losing real-time processing speed.
\end{itemize}

\section{Related works}

\noindent\textbf{Object detection}\bigskip

Detection frameworks have been dominated by sliding-window paradigm for many years. These methods heavily relied on hand-crafted features such as HOG\cite{dalal2005histograms}. However, after the dramatic performance boost brought on by R-CNN\cite{girshick2014rich}, which combines an object proposal mechanism\cite{uijlings2013selective} with a powerful CNN classifer, traditional methods were surpassed in a short period time. The R-CNN detector has been improved over the years both in terms of speed and accuracy. Recently, Faster-RCNN\cite{ren2015faster} integrated proposal generation module and the Fast R-CNN\cite{girshick2015fast} classifier into a single CNN. Many researchers adopted\cite{ren2015faster} framework and proposed a numerous extensions. This two-stage approaches consistently have occupied the top entries of the challenging benchmarks so far. However, due to the \textit{propose and classify} two-stage design, two-stage detector hurts the detection efficiency. They suffer from high memory usage and slow inference time. This motivates to build one-stage detectors that predicts outputs in a proposal-free manner.

OverFeat\cite{sermanet2013overfeat} is the first CNN based one-stage object detector using sliding-window paradigm. YOLO\cite{redmon2016you} and SSD \cite{liu2016ssd} have recently been proposed for real-time detection. They are a fast single stage methods which divide an image in to a multiple grids and simultaneously predict bounding boxes and class confidences. Unlike YOLO, SSD uses in-network multiple feature maps to detect objects with sizes of a specified ranges. This makes SSD more robust to detect varying shapes and sizes of objects than YOLO. We adopt the SSD framework for our starting point.

\bigskip\noindent\textbf{Using multiple layers and context information}\bigskip

A number of studies have shown that exploiting multiple layers within a CNN can improve detection and segmentation. HyperNet\cite{kong2016hypernet} and ION\cite{bell2016inside} concatenate the features from different layers and pool object proposals from the coupled layer. FCN\cite{long2015fully} and Hypercolumns\cite{hariharan2015hypercolumns} upsample multiple layers and combine partial scores of each layers for final decision. SSD\cite{liu2016ssd} enforces each layer to predict certain scale of objects by distributing various scales of default boxes to multiple layers. Similar to SSD, MS-CNN\cite{cai2016unified} also uses multiple feature maps for prediction and they newly introduced deconvolution layer to increase the resolution of feature maps. FPN\cite{lin2017feature} attempted to leverage the pyramidal shape of CNN. They augmented the CNN to build strong semantics at all scales of feature maps, enabled by nearest neighbor upsampling and lateral connections.

Global contexts are well known to play critical role in visual recognition problems. Recent architectures attempt to use this strong semantics for their specific tasks. 
DPM\cite{felzenszwalb2010object} integrated a global root model and finer local part models to represent deformable objects efficiently. Viewpoints and Keypoints\cite{tulsiani2015viewpoints} leverages the global viewpoint estimation to improve local keypoint predictions. RRC\cite{ren2017accurate} transfers each feature semantic information to other layers by stacking pooling and deconvolution layers upon SSD. \cite{noh2015learning, newell2016stacked,fu2017dssd} have shown that encoder-decoder, hourglass shape, is effective to propagate the context information. FPN\cite{lin2017feature} builds rich semantics at all levels by combining each layers. CoupleNet\cite{zhu2017couplenet} introduces global FCN branch to extract global semantics. All of which show that effective combination of the strong semantics(e.g. global context information) and fine local details improve the discrimination performance. Inspired by recent works, we propose to use top-down feature combining module to diffuse out the semantics effectively. Our proposed StairNet follows the SSD-style pyramid and thus it inherits the advantages of SSD, while produces more accurate models. We show that our model is simple and effective which outperforms current state-of-the art one-stage detectors.

\begin{figure*}
\begin{center}
 \includegraphics[width=1\linewidth]{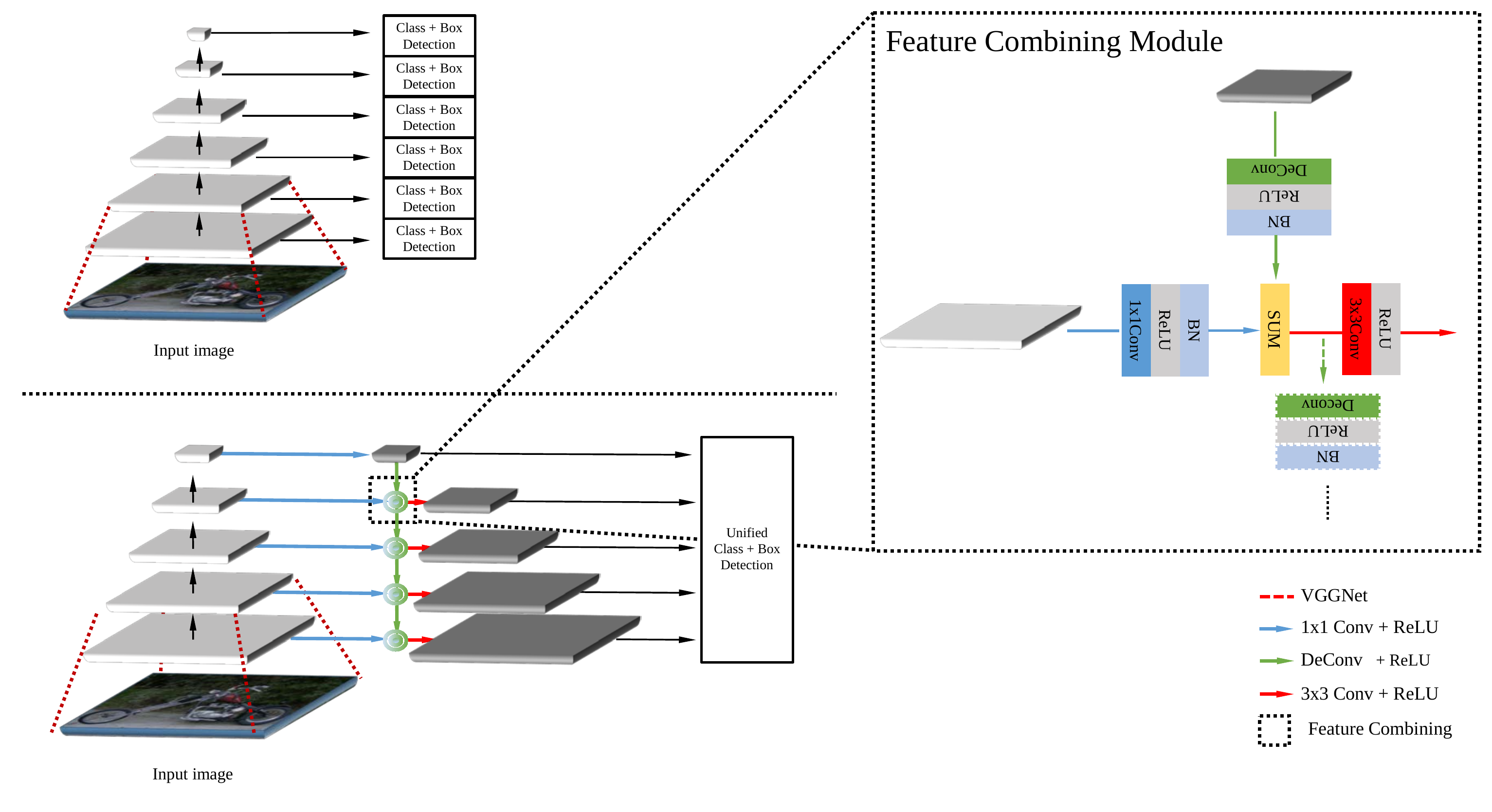}
\end{center}
  \caption{{\bf StairNet} architecture augments the SSD with {\bf feature combinining module}. The black dotted box shows the detail architecture of the feature combining module. This module spreads out the contextual information through down the layer effectively.}
\label{fig:StairNet}
\end{figure*}

\section{StairNet}

In this section, we begin by explaining the defect of the current detectors in details, then we elaborate the new improved detection framework, StairNet. We discuss why we choose the particular architecture and how we come up with our combining strategy.

\subsection{A weak spot of current detectors}

While recent CNN models designed for object detection have shown excellent capability to address the multiclass problem, less improvement has been made towards the detection of objects at various scales. For example, the Faster-RCNN\cite{ren2015faster} incorporated the proposal mechanism into single CNN. This saved computation time and enabled end-to-end training. However, for the object proposals, this approach only relies on the large receptive field of feature map(e.g. conv5). Since filter receptive fields are fixed but objects scales vary in natural scenes, this creates a discordance and compromises the performance. We can summarize this in a simple mathematical expression,

\begin{equation}
f_n = C_n(f_{n-1}) = C_n(C_{n-1}(...C_1(I))),
\label{eq:first}
\end{equation}

\begin{equation}
Object \ Proposals = P(f_n),
\label{eq:second}
\end{equation}

\noindent where {\bf$I$} is an input image, {\bf$C_n$} is a $n_{th}$ convolution block that is composed of convolutional layers, pooling layers, ReLU layers, etc. {\bf$f_n$} is the $n_{th}$ layer feature map, {\bf$P$} is the prediction layer that transforms certain feature map to detection ouputs: class confidence score and bounding box location. 

Recently in order to resolve the problem of \eref{eq:second}, SSD\cite{liu2016ssd} and MS-CNN\cite{cai2016unified} focused on the fact that the internal feature maps of a deep CNN are already of multi-scale, pyramidal shape. They utilizes the low-resolution maps to detect large objects and high-resolution map to detect small objects. These two approaches can be expressed as follows,


\begin{align}
\begin{split}
Detection \ Outputs & = \{P_{n-k}(f_{n-k}), ..., P_{n}(f_n)\},\\
                    & where \ n > k > 0
\end{split}
\label{eq:third}
\end{align}

\noindent Since they directly enforce each layer to be responsible for certain scale, every layers that are used for prediction have to be semantically strong.

It is well known that SSD-style detectors have inferior performance on small objects while they are competitive with state-of-the-art two-stage detectors on large objects.\cite{huang2017speed} We conjecture that this is because the lower level feature maps do not contain strong semantics(e.g contextual information). Due to different receptive field sizes of each feature maps, they differ in the level of semantics they are containing. In other words, as the feature map level goes down, semantic level gradually decreases.($n \rightarrow n-k$) The lowest layer then contains weak semantics, local features. \cite{zhou2015object} found that the actual receptive field\textbf{(arf)} size is much smaller than the theoretical receptive field({\bf trf}) size. \cite{luo2016understanding} shows that the pixels near the center of receptive field have much larger effect than the outer pixels, leading to a 2D-gaussian shape that also occupies smaller fraction than the trf. These findings indicate that the trf size sets an upper bound on the arf size. Since the {\bf arf} size of $f_{n-k}$ that is responsible for small objects in SSD is 58.6\cite{xiang2017context}, we can infer that $f_{n-k}$ seriously lacks global context information and only sees the local part of image that has size of 300 or 512. In \eref{eq:third}, it does not considers this problem well and directly uses $f_{n-k}$ to detect small objects, leading to poor performance. Therefore, we propose a new effective expression to handle this problem as follow,


\begin{align}
\begin{split}
Detection & \ Outputs = \{P_{n-k}(f'_{n-k}), ..., P_{n}(f'_n)\} \\
f'_n        & = f_n \\
f'_{n-1}    & = f_n + f_{n-1} \\
\vdots \\
f'_{n-k}    & = f_n + f_{n-1} + ... + f_{n-k}, \\
            & where \ n > k > 0
\end{split}
\label{eq:fourth}
\end{align}

The \eref{eq:fourth} differs from \eref{eq:second} and \eref{eq:third}, in that it uses multi-scale representations and distributes the stronger semantics gradually staring from the top layer. This gradual semantic aggregation is enabled by our iterative top-down feature combining operation. One thing to note that is still a one-stage process.


\subsection{NetworkArchitecture}

 The proposed StairNet is a single, unified network composed of 1) a meta-architecture(SSD), 2) the feature combining module, and 3) an unified prediction layer. \figref{fig:StairNet} illustrates the architecture of proposed \textbf{StairNet} framework along with the original SSD. Our framework first processes an arbitrary single-scale image in a fully convolutional way. Then the feature combining module distributes the semantics through down the layer, starting from the top-most feature map that generally contains strong semantics. The enhanced multi-scale feature maps are then referenced by the unified classifier to output the final predictions. We show that our StairNet improves the abstraction level of lower layer effectively. We elaborate each component in the following.

\subsubsection{\bf Meta-architecture of StairNet}\indent

Our first design principle is to leverage an in-network feature pyramid, hence, we adopt the SSD framework as the (meta)-architecture of StairNet. In particular, within the SSD, the feature extractor can be substituted for recent off-the-shelf CNNs\cite{he2016deep, huang2017densely, szegedy2017inception}. However in order to fairly measure the effectiveness of our proposed feature combining module, we maintained the feature extractor \cite{simonyan2014very} identical to the original SSD.(\figref{fig:StairNet}) 

Since the SSD contains multi-scale feature maps with a scaling step of 2, we take total five scales of feature maps that have strides of \{8, 16, 32, 64, 100\} pixels with respect to the input images. We take first two of feature maps from the base network(conv4\_3 and conv5\_2 from VGGNet). Then the remaining three feature maps are selected from the output of the two-stride subsampling layers that are added after the base network. 
We adopt different default box scale distribution from the original SSD. Given five level of feature maps, we set the scale of default boxes to be \{0.1, 0.2, 0.37, 0.54, 0.71\} respectively. We used the aspect ratio of default boxes to be \{2, 3\} in all scales. 



\subsubsection{\bf Feature Combining Module}\indent

Our second design principle is to make all feature maps semantically strong. We augment the conventional SSD with our feature combining module in order to propagate the high-level abstraction features of top layer to lower layer. (\figref{fig:StairNet} black dotted box) illustrates the unit of feature combining module. It consists of three parts: 1x1 convolution layer, deconvolution layer and 3x3 convolution layer.

In order to combine the propagated information of upper layer and the original features of corresponding bottom layer, we introduced \textbf{1x1 convolution layers}. We adopt whole channels of feature maps to be 256 which is the minimum of original values\{512, 1024, 512, 256, 256\}.(\figref{fig:StairNet} blue line) This choice is natural since it allows similar levels of influence when two feature maps are combined.
To combine two different size of feature maps, we added \textbf{deconvolution layers} that upsample by a factor of 2.(\figref{fig:StairNet} green line) The features of upper layer, that have more strong semantics relative to lower layer, are delivered by this deconvolution layer.
Before combining them together, it is essential to normalize features from different layers since it shows different scale distribution.\cite{liu2015parsenet} Note that we used batch normalization to handle this problem.
We then combine them using element-wise add operation.
The combined features are passed down directly to the next deconvolution layer. The spectrum of information sources of each feature maps incrementally increases by this iterative process. The lower the feature map, the more the information sources are supplied to complement weak semantics of lower feature maps.
To effectively mix the information from different streams(\figref{fig:StairNet} blue and green line), we applied a \textbf{3x3 convolution layer}(\figref{fig:StairNet} red line) to construct the final enhanced feature maps before the classifier. Note that the enhanced feature maps have same spatial sizes respect to the original feature maps.

\subsubsection{\bf Unified Prediction Layer}\indent

As shown in \tabref{table:ablation} col7 and col8, despite of reducing the number of parameters by employing unified (shared) classifier, performance remains the same. Many previous works argued that object scale difference causes different distribution in feature space \cite{junjie2013mtdpm, jianan2015scaleaware}. Thus, like SSD\cite{liu2016ssd}, multiple classifiers for different scales have been popular choice for better performance. In this point of view, the result in \tabref{table:ablation} implies that proposed feature combining module effectively mitigate the semantic gap between different hierarchical layers, i.e. the feature maps share similar degree of semantics in spite of scale difference. In this reason, by adopting the unified classifier we can reduce the parameters of classifier. Potential advantage of similar semantic representation over hierarchical layers and unified classifier is alleviation of training data imbalance problem over object scales. For the skewed distribution of a specific category, e.g. there are many large cows but few small cows in PASCAL VOC 2007, our method could be helpful because our module shares a single classifier over various scales: as the final feature representations of large cow and small cow are similar, the classifier trained with large cows would work well for small cows as well.


Like the other one-stage classifiers, our unified prediction layer predicts the probability of object presence and bounding box offsets, at each spatial location for each of the $k$ default boxes and $c$ object classes. Taking the outputs of feature combing module as input, this prediction layer applies a 3x3 conv layer with $(c+4)k$ filters.

\begin{table}
\centering
\resizebox{\columnwidth}{!}{%
\begin{tabular}{c c c c}
\hline
     Extra ConvBlock & {\bf SUM} &  MAX &PRODUCT \\
\hline
    {\bf w 3x3Conv} & {\bf 78.8 (76.4)}     & 78.9 (76.2)   & 78.3\\
    w/o 3x3Conv     & 78.2                  & 77.8          & -   \\
\hline
\end{tabular}
}
\smallskip
\caption{{\bf Effects of Extra ConvBlock and Combining Methods} Metric: detection mAP(\%) on VOC07 \textit{test}. ( \ ): detection results of PASCAL VOC12 \textit{test}.}
\label{table:ele_op}
\end{table} 

\begin{table}
\centering
\resizebox{\columnwidth}{!}{%
\begin{tabular}{c |c c c |c c c c |c}
                    & SSD     &           &           &           &           &           &           & {\bf StairNet}\\
\hline
    \{2, 3\}        &         &\checkmark &\checkmark &           &\checkmark &\checkmark &\checkmark &\checkmark \\
    deconv          &         &           &\checkmark &\checkmark &           &\checkmark &\checkmark &\checkmark\\
    3x3 conv        &         &           &           &\checkmark &\checkmark &           &\checkmark &\checkmark\\
    unified         &         &\checkmark &\checkmark &\checkmark &\checkmark &\checkmark &           & \checkmark \\
\hline
    \{1.6, 2, 3\}   &         &           &           &\checkmark &           &           &\\
    bilinear        &         &           &           &           &\checkmark &           &\\
    ResBlock        &         &           &           &           &           &\checkmark &\\
    multi           &         &           &           &           &           &           &\checkmark &\\
\hline
    VOC 2007 mAP    & 77.2    & 77.4      & 78.2      & 78.9      & 78.5      & 78.6      & 78.9      & {\bf 78.8}\\
\end{tabular}
}
\smallskip
\caption{{\bf Ablation Experiments on StairNet.} Each component of the first row corresponds to each component of the second row in an one-to-one manner. {\bf \{2,3\} and \{1.6, 2, 3\}} indicates the aspect ratio of the default boxes. {\bf deconv and bilinear} indicates the upsampling method in the feature combining module. {\bf 3x3 conv and ResBlock} indicates the extra layers in the feature combining module. {\bf unified and multi} indicates whether the prediction layer shares the weight or not.}
\label{table:ablation}
\end{table}

\begin{table*}
\resizebox{\textwidth}{!}{\begin{tabular}{c| c| c| c | c c c c c c c c c c c c c c c c c c c c c }
\hline
Method & data & network & map & aero & bike & bird & boat & bottle & bus & car & cat & chair & cow & table & dog & horse & mbike & persn & plant & sheep & sofa & train & tv\\
\hline
HyperNet \cite{kong2016hypernet}            & 07+12   & VGGNet       & 76.3
    & 77.4 & 83.3 & 75.0 & 69.1 & 62.4 
    & 83.1 & 87.4 & 87.4 & 57.1 & 79.8 
    & 71.4 & 85.1 & 85.1 & 80.0 & 79.1 
    & 51.2 & 79.1 & 75.7 & 80.9 & 76.5 \\
    
Fr R-CNN \cite{ren2015faster}           & 07+12   & ResNet-101   & 76.4 
    & 79.8 & 80.7 & 76.2 & 68.3 & 55.9 
    & 85.1 & 85.3 & 89.8 & 56.7 & 87.8 
    & 69.4 & 88.3 & 88.9 & 80.9 & 78.4 
    & 41.7 & 78.6 & 79.8 & 85.3 & 72.0 \\

ION \cite{bell2016inside}                   & 07+12+S & VGGNet       & 76.5 
    & 79.2 & 79.2 & 77.4 & 69.8 & 55.7 
    & 85.2 & 84.2 & 89.8 & 57.5 & 78.5 
    & 73.8 & 87.8 & 85.9 & 81.3 & 75.3 
    & 49.7 & 76.9 & 74.6 & 85.2 & 82.1 \\
    
R-FCN \cite{dai2016r}                       & 07+12   & ResNet-101   & 80.5 
    & 79.9 & 87.2 & 81.5 & 72.0 & 69.8 
    & 86.8 & 88.5 & 89.8 & 67.0 & 88.1 
    & 74.5 & 89.8 & 90.6 & 79.9 & 81.2 
    & 53.7 & 81.8 & 81.5 & 85.9 & 79.9 \\

\hline

{\bf SSD300*} \cite{liu2016ssd}             & 07+12   & VGGNet       & {\bf 77.2}         
    & 82.3 & 84.5 & 75.0 & 69.9 & 51.2
    & 85.2 & 85.6 & 87.5 & 63.0 & 82.6 
    & 76.2 & 84.2 & 86.5 & 83.8 & 78.6
    & 51.0 & 75.1 & 79.6 & 86.7 & 75.5 \\ 
    
SSD300 \cite{liu2016ssd}                    & 07+12   & VGGNet       & 77.5 
    & 79.5 & 83.9 & 76.0 & 69.6 & 50.5
    & 87.0 & 85.7 & 88.1 & 60.3 & 81.5 
    & 77.0 & 86.1 & 87.5 & 83.9 & 79.4 
    & 52.3 & 77.9 & 79.5 & 87.6 & 76.8 \\
    
DSSD321 \cite{fu2017dssd}                  & 07+12   & ResNet-101   & 78.6 
    & 81.9 & 84.9 & 80.5 & 68.4 & 53.9
    & 85.6 & 86.2 & 88.9 & 61.1 & 83.5 
    & 78.7 & 86.7 & 88.7 & 86.7 & 79.7 
    & 51.7 & 78.0 & 80.9 & 87.2 & 79.4 \\

\hline

{\bf StairNet}      & 07+12   & VGGNet       & {\bf 78.8}     
    & 81.3 & 85.4 & 77.8 & 72.1 & 59.2
    & 86.4 & 86.8 & 87.5 & 62.7 & 85.7
    & 76.0 & 84.1 & 88.4 & 86.1 & 78.8
    & 54.8 & 77.4 & 79.0 & 88.3 & 79.2 \\
\end{tabular}}
\begin{tablenotes}
\scriptsize
\item \hspace*{\fill} {\bf *}: reproduced in PyTorch framework.
\end{tablenotes}
\caption{{\bf PASCAL VOC 2007 \textit{test} detection results.} {\bf 07+12}: 07 trainval + 12 trainval. {\bf 07+12+S}: 07+12 plus segmentation labels.}   
\label{table:voc07}
\end{table*}

\begin{table*}
\resizebox{\textwidth}{!}{\begin{tabular}{c| c| c| c | c c c c c c c c c c c c c c c c c c c c c }
\hline
Method & data & network & mAP & aero & bike & bird & boat & bottle & bus & car & cat & chair & cow & table & dog & horse & mbike & persn & plant & sheep & sofa & train & tv\\ 
\hline
HyperNet \cite{kong2016hypernet}            & 07++12    & VGGNet       & 71.4 
    & 84.2 & 78.5 & 73.6 & 55.6 & 53.7
    & 78.7 & 79.8 & 87.7 & 49.6 & 74.9
    & 52.1 & 86.0 & 81.7 & 83.3 & 81.8
    & 48.6 & 73.5 & 59.4 & 79.9 & 65.7 \\
Fr R-CNN \cite{ren2015faster}           & 07++12    & ResNet-101   & 73.8 
    & 86.5 & 81.6 & 77.2 & 58.0 & 51.0
    & 78.6 & 76.6 & 93.2 & 48.6 & 80.4
    & 59.0 & 92.1 & 85.3 & 84.8 & 80.7
    & 48.1 & 77.3 & 66.5 & 84.7 & 65.6 \\
ION \cite{bell2016inside}                   & 07++12+S  & VGGNet       & 76.4     
    & 87.5 & 84.7 & 76.8 & 63.8 & 58.3
    & 82.6 & 79.0 & 90.9 & 57.8 & 82.0
    & 64.7 & 88.9 & 86.5 & 84.7 & 82.3
    & 51.4 & 78.2 & 69.2 & 85.2 & 73.5 \\
RFCN m-sc \cite{dai2016r}               & 07++12    & ResNet-101   & 77.6 
    & 86.9 & 83.4 & 81.5 & 63.8 & 62.4
    & 81.6 & 81.1 & 93.1 & 58.0 & 83.8
    & 60.8 & 92.7 & 86.0 & 84.6 & 84.4
    & 59.0 & 80.8 & 68.6 & 86.1 & 72.9 \\
\hline
YOLOv2 \cite{redmon2017yolo9000}            & 07++12    & Darknet-19   & 73.4 
    & 86.3 & 82.0 & 74.8 & 59.2 & 51.8 
    & 79.8 & 76.5 & 90.6 & 52.1 & 78.2 
    & 58.5 & 89.3 & 82.5 & 83.4 & 81.3
    & 49.1 & 77.2 & 62.4 & 83.8 & 68.7 \\ 
{\bf SSD300*} \cite{liu2016ssd}             & 07++12    & VGGNet       & {\bf 74.8} 
    & 87.7 & 83.4 & 73.0 & 60.3 & 47.7 
    & 80.6 & 76.7 & 91.5 & 57.6 & 77.6
    & 63.6 & 89.0 & 84.9 & 84.8 & 81.8
    & 48.9 & 77.9 & 72.3 & 86.1 & 71.2 \\ 
SSD300 \cite{liu2016ssd}                    & 07++12    & VGGNet       & 75.8 
    & 88.1 & 82.9 & 74.4 & 61.9 & 47.6 
    & 82.7 & 78.8 & 91.5 & 58.1 & 80.0
    & 64.1 & 89.4 & 85.7 & 85.5 & 82.6
    & 50.2 & 79.8 & 73.6 & 86.6 & 72.1 \\
DSSD321 \cite{fu2017dssd}                   & 07++12    & ResNet-101   & 76.3 
    & 87.3 & 83.3 & 75.4 & 64.6 & 46.8
    & 82.7 & 76.5 & 92.9 & 59.4 & 78.3
    & 64.3 & 91.5 & 86.6 & 86.6 & 82.1
    & 53.3 & 79.6 & 75.7 & 85.2 & 73.9 \\
\hline
{\bf StairNet}      & 07++12    & VGGNet       & {\bf 76.4} 
    & 87.7 & 83.1 & 74.6 & 64.2 & 51.3
    & 83.6 & 78.0 & 92.0 & 58.9 & 81.8
    & 66.2 & 89.6 & 86.0 & 84.9 & 82.6
    & 50.9 & 80.5 & 71.8 & 86.2 & 73.5 \\
\end{tabular}}
\begin{tablenotes}
\scriptsize
\item \hspace*{\fill} {\bf *}: reproduced in PyTorch framework.
\end{tablenotes}
\caption{{\bf PASCAL VOC 2012 \textit{test} detection results.} {\bf 07++12}: 07 trainval + 07 test + 12 trainval. {\bf 07+12+S}: 07+12 plus segmentation labels. Result link of \textbf{StairNet} : \url{http://host.robots.ox.ac.uk:8080/anonymous/SPPVPF.html} } 
\label{table:voc12}
\end{table*}

\begin{table*}
\resizebox{\textwidth}{!}{\begin{tabular}{c| c| c | c c c c c c c c c c c c c c c c c c c c c }
\hline
Method & scale & mAP & aero & bike & bird & boat & bottle & bus & car & cat & chair & cow & table & dog & horse & mbike & persn & plant & sheep & sofa & train & tv \\ 
\hline
SSD300* \cite{liu2016ssd}
    & small     & 41.7         
        & 47.2 & 64.7 & 35.2 & 23.2 &  8.5    
        & 45.7 & 49.6 & 67.2 & 20.0 & 47.6      
        & 30.1 & 53.1 & 61.6 & 46.9 & 27.8    
        &  7.3 & 44.5 & 62.0 & 63.3 & 28.3 \\
{\bf StairNet}
    & small     & \textbf{50.6}
        & 57.5 & 71.6 & 51.4 & 34.5 & 21.8
        & 54.2 & 56.2 & 65.4 & 24.0 & 73.2
        & 31.6 & 60.4 & 73.3 & 54.1 & 36.7
        & 21.5 & 51.7 & 59.5 & 70.0 & 43.8 \\
\hline
SSD300* \cite{liu2016ssd}
    & medium    & 76.8
        & 79.9 & 81.6 & 75.6 & 65.0 & 44.4    
        & 88.0 & 86.5 & 87.3 & 66.2 & 83.5
        & 74.9 & 83.6 & 87.7 & 86.5 & 80.3    
        & 46.5 & 75.2 & 78.9 & 86.1 & 77.0 \\
{\bf StairNet}
    & medium    & {\bf 78.0}
        & 79.1 & 83.1 & 77.9 & 69.2 & 55.0    
        & 88.1 & 86.7 & 88.0 & 62.5 & 83.1
        & 78.0 & 83.1 & 89.3 & 88.5 & 79.1    
        & 49.7 & 74.5 & 79.4 & 87.7 & 78.5 \\
\hline
SSD300* \cite{liu2016ssd}
    & large     & {\bf 80.4}
        & 90.5 & 88.1 & 80.1 & 80.5 & 63.7    
        & 88.3 & 90.2 & 90.7 & 50.9 & 82.6
        & 84.9 & 84.8 & 87.8 & 90.1 & 85.5    
        & 57.7 & 82.1 & 62.4 & 88.3 & 78.8 \\
{\bf StairNet}
    & large     & 78.5
        & 90.4 & 89.3 & 85.8 & 76.2 & 63.5    
        & 92.0 & 89.4 & 88.4 & 47.2 & 81.8
        & 80.3 & 80.5 & 88.2 & 89.8 & 83.5    
        & 52.8 & 76.6 & 46.5 & 89.9 & 79.0 \\
\end{tabular}}
\begin{tablenotes}
\scriptsize
\item \hspace*{\fill} {\bf *}: reproduced in PyTorch framework.
\end{tablenotes}
\caption{ {\bf Scale-aware evaluation on PASCAL VOC 2007 \textit{test}.} Two methods are trained on VOC 07+12. VGGNet is used as backbone in both methods.} 
\label{table:scale}
\end{table*}

~
~ 
~

\section{Experiments}

We evaluate the StairNet on the widely used datasets: PASCAL VOC 2007 and VOC 2012 benchmarks\cite{everingham2010pascal}. All of our experiments are based on the Pytorch framework. In order to better perform apple-to-apple comparisons, we first attempted to reproduce the original accuracy of SSD in the Pytorch framework and set as our baseline. Then we performed extensive ablation studies to thoroughly investigate the effectiveness of each component. Moreover, we evaluate our model on different object scales and verify that StairNet improvoes the weakness of SSD. Finally, we show that StairNet outperforms current state-of-the-art one-stage detectors.
 
\subsection{Ablation studies on VOC2007}

We perform ablation experiments on PASCAL VOC 2007 test sets for detailed analysis of our proposed StairNet framework. We train the models on the union set of VOC 2007 trainval and VOC2012 trainval (“07+12”), and evaluate on VOC 2007 test set. We first removed each component step-by-step to observe real effects of each component.(\tabref{table:ablation} cols 9,4,3 and 2) We then conduct controlled experiment to investigate several design choices in our model.(\tabref{table:ablation} cols 9,8,7,6 and 5)) In all the experiments, the size of input image is fixed to 300 for simplicity. The results are mainly summarized in \tabref{table:ele_op} and \tabref{table:ablation}.

\subsubsection{Combining methods: Element-wise operation}

\smallskip\noindent\textbf{Element-wise sum}\smallskip

We fist investigate three different combining methods: element-wise sum, element-wise product and element-wise max. (\tabref{table:ele_op} row 2) shows that using element-wise sum generates the best performance. This phenomenon can be interpreted in terms of information flow. The ResidualNet\cite{he2016deep}, which achieves state of the art results in many challenging vision tasks, shows that the element-wise sum is effective way to integrate and preserve the information. In forward phase, it enables network to use the information from two branches complementary without losing any of information. In the backward phase, the gradient is distributed equally to all of inputs, leading to an efficient training. The element-wise maximum, which routes the gradient only to the higher inputs provides regularization effect in some extent, it generates unstable performance. The element-wise product, which assigns a small gradient to the large input and a large gradient to the small input leads network hard to converge, yielding the worst performance. Therefore, we use element-wise sum for the following experiments.

\subsubsection{Analysis of each component in StairNet}
\label{sec:extra3x3conv}

\smallskip\noindent\textbf{Extra ConvBlock}\smallskip

As shown in \tabref{table:ablation} of col 4, we observe performance drop without the 3x3 convolution layer. In order to investigate the role of 3x3 convolution layer thoroughly, we conducted additional experiment shown in \tabref{table:ele_op}. We observe that without the 3x3 convolution layer, performances are consistently degraded regardless of the type of combining methods. In the case of element-wise product, the training was even not stable without the 3x3 convolution layer. This experiment shows that the extra 3x3 convolution layer not only improves the performance but also helps training more stable. Since introducing 3x3 convolution layer makes the total depth of the model more deep, it increases the capacity of the model. Moreover, we conjecture that 3x3 convolution layer acts like a buffer that constructs similar semantic levels of final feature maps before the unified classifier.

\bigskip\noindent\textbf{Top-bottom connection}\smallskip

As shown in \tabref{table:ablation} of col 3, we observe significant drop of performance without the deconvolution layer, i.e. removing top-bottom connection. This shows that top-down semantic aggregation plays a critical role in feature combining module. The deconvolution layer helps the distribution of semantics from top to bottom and reduces the semantic gap of feature maps. We can also adopt naive upsampling methods such as nearest neighbor and bilinear interpolation.\cite{lin2017feature} However we found deconvolution layer performs better than simple upsampling method. We will discuss this point in the following section.



\begin{table}
\centering
\resizebox{\columnwidth}{!}{%
\begin{tabular}{c|c|c|c|c|c}
\hline
    Method & data & network & mAP & fps & lib\\
\hline
    YOLO \cite{redmon2017yolo9000}          & 07+12     & GoogLeNet  & 63.4        & 45    & DarkNet\\
    YOLOV2\_352 \cite{redmon2017yolo9000}   & 07+12     & DarkNet-19 & 73.7        & 81    & DarkNet\\
    YOLOV2\_544 \cite{redmon2017yolo9000}   & 07+12     & DarkNet-19 & 78.6        & 40    & DarkNet\\
    \rowcolor{Gray}
    SSD300*\cite{liu2016ssd}                & 07+12     & VGGNet     & 77.2        & 42    & PyTorch\\
    \rowcolor{Gray}
    SSD300 \cite{liu2016ssd}                & 07+12     & VGGNet     & 77.5        & 62    & Caffe  \\
    DSSD321 \cite{fu2017dssd}               & 07+12     & ResNet-101 & 78.6        & 9.5   & Caffe  \\
    RSSD300 \cite{jisoo2017rssd}            & 07+12     & VGGNet     & 78.5        & 35.0  & Caffe  \\
    DiCSSD300\cite{xiang2017context}        & 07+12     & VGGNet     & 78.1        & 40.8  & Caffe \\
\hline
    {\bf StairNet}                          & 07+12     & VGGNet     & {\bf 78.8}  & 30    & PyTorch\\
\end{tabular}}
\begin{tablenotes}
\scriptsize
\item \hspace*{\fill} {\bf *}: reproduced in PyTorch framework.
\end{tablenotes}
\caption{{\bf PASCAL VOC 2007 \textit{test} detection results.} Is is worth to note that PyTorch implementation runs slower (62fps $\rightarrow$ 42fps) and shows lower performance (77.5$\rightarrow$77.2) than Caffe implementation for exactly same algorithm (Row 4 \& 5). In spite of this disadvantage, \textbf{StairNet} outperforms other state-of-the-art methods.} 
\label{table:sota_voc07}
\end{table} 

\begin{table}
\centering
\resizebox{\columnwidth}{!}{%
\begin{tabular}{c| c |c c c c c c|c}
\hline
&&&&Recall&&&&\\
Method & data&0.5&0.6&0.7&0.8&0.9&1&mAP@0.7+  \\
\hline
    SSD300* \cite{liu2016ssd}    & 07+12 &91.9&87.9&79.7&65.6&34.4&0& 44.9  \\
    R-SSD300 \cite{jisoo2017rssd}& 07+12 &92.7&88.4&82.4&68.9&37.6&0& 47.2  \\
    \textbf{StairNet}            & 07+12 &\textbf{94.3}&\textbf{90.1}&\textbf{83.5}&\textbf{70.1}&\textbf{38.8}&0& \textbf{48.1}  \\
\hline
\end{tabular}}
\begin{tablenotes}
\scriptsize
\item \hspace*{\fill} {\bf *}: reproduced in PyTorch framework.
\end{tablenotes}
\smallskip
\caption{{\bf mAP at Recall$\ge$0.7} mean average precision over recall$\ge$0.7 suggested by \cite{jisoo2017rssd}. In most practical cases, it is more important to achieve high-precision at high-recall range rather than at low-recall range. Our \textbf{StairNet} outperforms SSD and R-SSD. }
\label{tab:map07}
\end{table} 


\subsubsection{Model architecture design}

\bigskip\noindent\textbf{\{2,3\} aspect ratio vs \{1.6, 2, 3\} aspect ratio}\smallskip

Recently \cite{fu2017dssd} conducted k-means clustering on the bounding boxes in the training data of PASCAL VOC 2007 and VOC 2012 and they included one more aspect ratio of 1.6 to improve the performance. However, as shown in (\tabref{table:ablation} cols 5 and 9) we observe no significant improvement with it. Since using less aspect ratio not only saves weight parameters but also computation times, we stick to \{2,3\} aspect ratio.

~
~

\bigskip\noindent\textbf{Top-bottom connection: Deconv vs Bilinear}\smallskip

Generally, there are two ways to enlarge the resolution(x2) of feature map. First, the deconvolution layer that their upsampling weights are learned through the training process. Second, naive upsampling methods such as nearest neighbor or bilinear interpolation. As can be seen in (\tabref{table:ablation} cols 6 and 9) deconvolution layer improves the performance. This implies that the learned upsampling-weights perform better than the naive upsampling kernels. Moreover, the recent studies\cite{noh2015learning, fu2017dssd} have shown that the sequence of deconvolution layer is suitable for propagating the information efficiently.

\bigskip\noindent\textbf{Extra ConvBlock: 3x3Conv vs ResBlock}\smallskip

 We already have shown the effectiveness of 3x3 convolution layer in \sref{sec:extra3x3conv}. Rather than 3x3 convolution layer, we experiment with ResBlock to make each detection path deeper. The detail architecture of ResBlock\footnote{Standard basic block: [(C1-R)+(C1-R-C3-R-C1)]-Sum-R} is explained in supplemental section. As shown in the (\tabref{table:ablation} cols 7 and 9), we observe that 3x3 convolution layer performs better than ResBlock. We conjecture that since the combined features which are fed into ResBlock would have redundant information, skip path in ResBlock would deliver unnecessary information which causes performance degradation.
 
\bigskip\noindent\textbf{Unified Classifier vs Multi Classifier}\smallskip

 Unlike original SSD, our final StairNet uses unified classifier. As shown in (\tabref{table:ablation} cols 8 and 9) we observe no big difference on performance which indicates that all feature maps share similar degree of semantics. This justifies the effectiveness of our feature combining module that spreads out the information effectively. We adopt unified classifier for our final model to save weight parameters.

\subsection{Impact on different sized objects}

To analyze the impact of our final model on the detection performance of different sized objects, we evaluated SSD and StairNet, considering objects of three different sizes. We find that the MS COCO criteria for object scales causes serious imbalance on VOC2007 test set, leading to undesirable comparison. We show this statistics in supplemental section. Instead, for each class we sorted ground truth bounding boxes on test set by area and divided them into three part: $small:[\ \sim25 \%)$, $medium:[25\%\sim 75\%)$ and $large:[75\%\sim \ ]$ which consistently results in 1:2:1 ratio of number of sizes of test set for each class. When benchmarking on objects of each size, the ground truth labels for other sizes were ignored. As shown in \tabref{table:scale}, proposed method shows significantly better performance than SSD on small scales.(8.9 mAP increase) StairNet wins on 18 classes among 20 classes. Even though the non-rigid objects such as cow, horse, person, and bird look very different due to its deformability, StairNet works better on these categories because it captures contextual information.

\subsection{PASCAL VOC 2007 Results}

We trained our model on the union of 2007 trainval and 2012 trainval. We used the same training scheme for both SSD and StairNet. We used a weight decay of 0.0001 and a momentum of 0.9. A batch size was set to 16 and adopted SGD optimizer with initial learning rate 0.001. We then decreased it by a factor of 10 at 80K and 100K iterations respectively. The training was terminated at 120K iterations. 
 
\tabref{table:voc07} shows our results on the PASCAL VOC 2007 test set. SSD* is the reproduced version in Pytorch framework by ourselves and we achieved 77.2 \%. StairNet achieves a mAP of 78.8 \%, which outperforms the SSD by 1.6 points. Our model even outperforms the DSSD which uses ResNet-101 as their base network. Note that our StairNet model shows a large improvement over the classes with specific backgrounds like boat, car, cow, train, i.e. water for boat and railroad for train and so on. We also observed significant gain over the objects that mainly contain a small sizes of ground truth boxes such as bottle and plant.
 
We evaluate the inference time of our network using a NVIDIA-TITAN X GPU (pascal) along with CUDA 8.0 and cuDNN-v5.1. As shown in \tabref{table:sota_voc07}, StairNet outperforms all the current one-stage methods in 30fps. One thing to note is that Pytorch implementation runs slower and shows lower perfomance than original Caffe implementation for exactly same algorithm. Inspite of this disadvantage, StairNet outperforms other state-of-the-art methods including most recent SSD-based detectors.\cite{fu2017dssd, jisoo2017rssd, xiang2017context}
 
Moreover we evaluate our model on mean average precision over Recall$\ge$0.7 suggested by \cite{jisoo2017rssd}. \tabref{tab:map07} shows the results that even in the high-recall range our model achieves high-precision and outperforms SSD and R-SSD\cite{jisoo2017rssd}.

\subsection{PASCAL VOC 2012 Results}

We also evaluate our method on the more challenging VOC2012 dataset by submitting results to the public evaluation server. We use VOC 2007 test, VOC 2007 trainval and VOC2012 trainval as the training set. We follow the same setting of VOC2007 except the number of total iterations. Since there are more training images we increased the number of training iterations to 140K. Starting from the same learning rate of 0.001, we then decreased it by a factor of 10 at 80k, 100k and 120k iterations respectively. The training was terminated at 140K iterations.

\tabref{table:voc12} shows the results on the VOC2012 test set. Our method achieves 76.4\% mAP, which outperforms the SSD by 1.6\%. As shown in the table we observe similar improvement over the specific class. The StairNet outperforms all other one-stage methods once again.


   


\begin{figure}
\centering
\hspace{-3em}
\subfigure[SSD300*]{
\begin{tabular}{c}
\includegraphics[width=40mm]{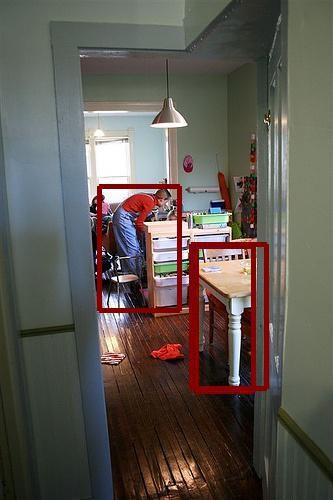} \\
\includegraphics[width=40mm]{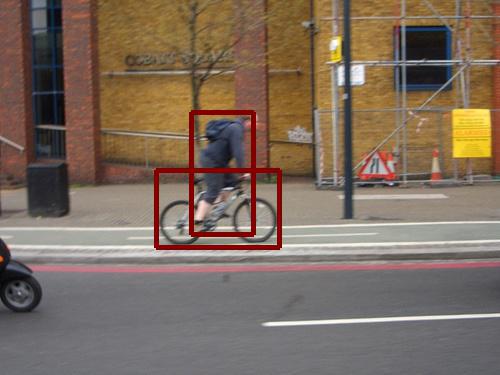} \\
\includegraphics[width=40mm]{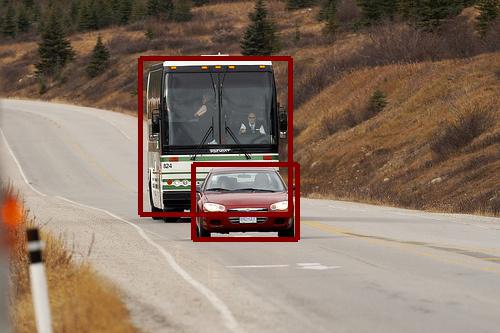} \\
\includegraphics[width=40mm]{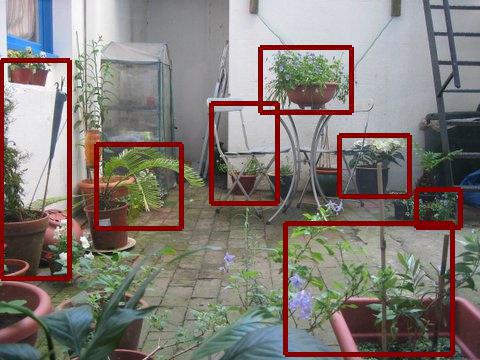} \\
\includegraphics[width=40mm]{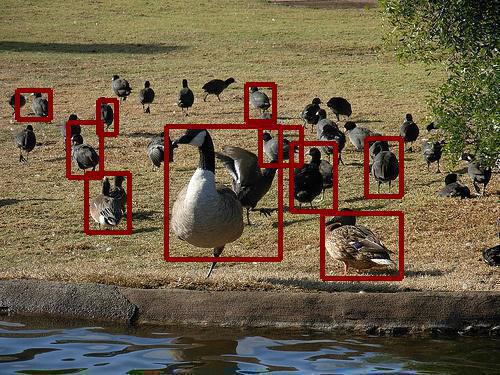} \\
\end{tabular}
}
\hspace{-2em}
\subfigure[StairNet300]{
\begin{tabular}{c}
\includegraphics[width=40mm]{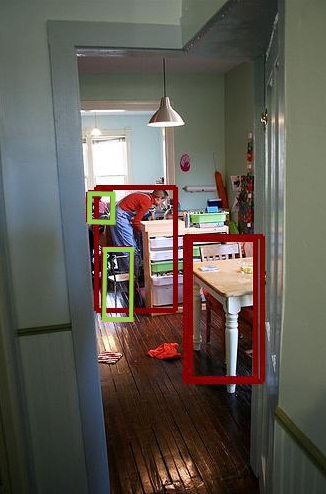} \\
\includegraphics[width=40mm]{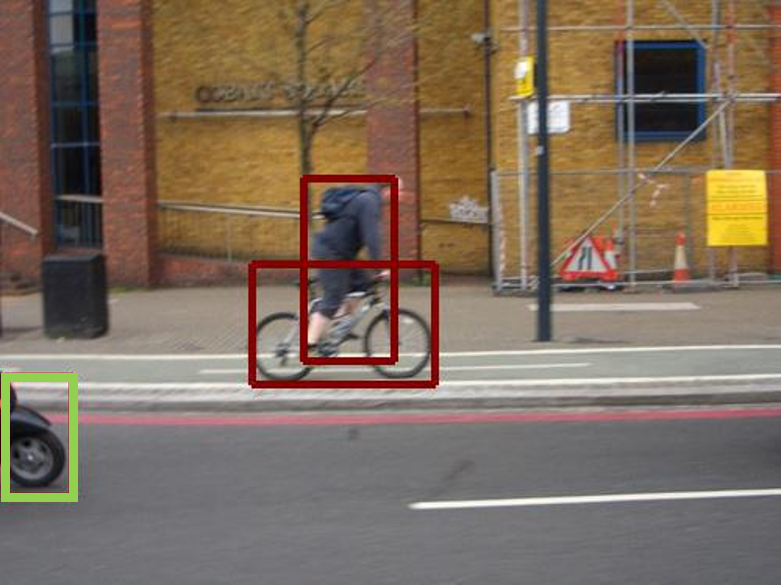} \\
\includegraphics[width=40mm]{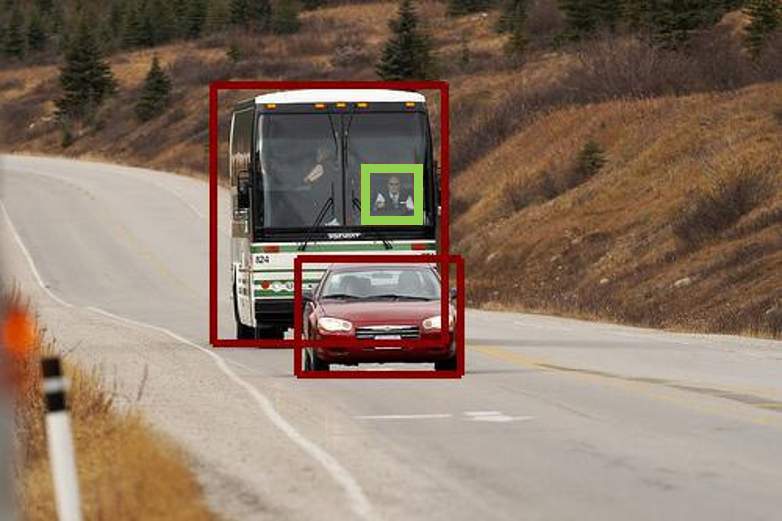} \\
\includegraphics[width=40mm]{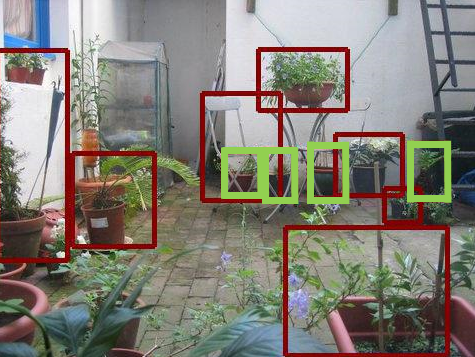} \\
\includegraphics[width=40mm]{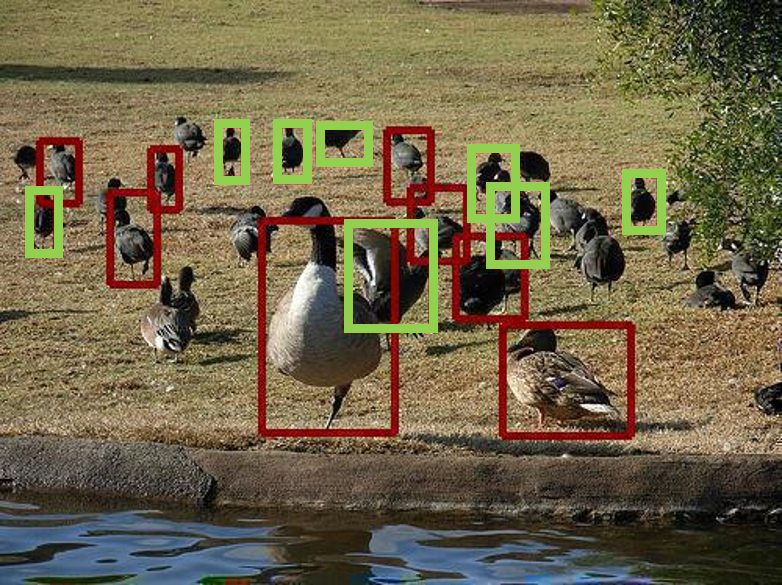} \\
\end{tabular}
}
\hspace{-3em}
\caption{\textbf{Qualitative results on PASCAL VOC 2007}. Green boxes indicate that \textbf{StairNet} performs better than SSD in challenge scenarios.}
\label{fig:results}
\end{figure}

\section{Conclusion}
In this paper, we present the StairNet an effective one-stage detector that spreads out the strong semantics in a top-down manner and constructs an enhanced multi-scale feature maps for accurate detection. We point out that two-stage methods do not utilize the advantages of inherent multi-scale feature maps while one-stage methods ignore to incorporate global context information. To address this, we augment the SSD framework with our feature combing module, leading to significant improvement on detecting small objects. The StairNet is simple and fast. We verify its efficacy by showing that it achieves state-of-the-art accuracy on two standartd benchmarks.

{\small
\bibliographystyle{ieee}
\bibliography{egpaper_final}

\begin{thebibliography}{10}\itemsep=-1pt

\bibitem{bell2016inside}
S.~Bell, C.~Lawrence~Zitnick, K.~Bala, and R.~Girshick.
\newblock Inside-outside net: Detecting objects in context with skip pooling
  and recurrent neural networks.
\newblock In {\em {Proc. of Computer Vision and Pattern Recognition (CVPR)}},
  2016.

\bibitem{cai2016unified}
Z.~Cai, Q.~Fan, R.~S. Feris, and N.~Vasconcelos.
\newblock A unified multi-scale deep convolutional neural network for fast
  object detection.
\newblock In {\em {Proc. of European Conf. on Computer Vision (ECCV)}}, 2016.

\bibitem{chen15deeplab}
L.-C. Chen, G.~Papandreou, I.~Kokkinos, K.~Murphy, and A.~L. Yuille.
\newblock Semantic image segmentation with deep convolutional nets and fully
  connected crfs.
\newblock In {\em {Proc. of Int'l Conf. on Learning Representations (ICLR)}},
  2015.

\bibitem{dai2016r}
J.~Dai, Y.~Li, K.~He, and J.~Sun.
\newblock R-fcn: Object detection via region-based fully convolutional
  networks.
\newblock In {\em {Proc. of Neural Information Processing Systems (NIPS)}},
  2016.

\bibitem{dalal2005histograms}
N.~Dalal and B.~Triggs.
\newblock Histograms of oriented gradients for human detection.
\newblock In {\em {Proc. of Computer Vision and Pattern Recognition (CVPR)}},
  2005.

\bibitem{everingham2010pascal}
M.~Everingham, L.~Van~Gool, C.~K. Williams, J.~Winn, and A.~Zisserman.
\newblock The pascal visual object classes (voc) challenge.
\newblock In {\em {Int'l Journal of Computer Vision (IJCV)}}, volume~88, pages
  303--338, 2010.

\bibitem{felzenszwalb2010object}
P.~F. Felzenszwalb, R.~B. Girshick, D.~McAllester, and D.~Ramanan.
\newblock Object detection with discriminatively trained part-based models.
\newblock In {\em {IEEE Trans. Pattern Anal. Mach. Intell. (TPAMI)}},
  volume~32, pages 1627--1645, 2010.

\bibitem{fu2017dssd}
C.-Y. Fu, W.~Liu, A.~Ranga, A.~Tyagi, and A.~C. Berg.
\newblock Dssd: Deconvolutional single shot detector.
\newblock {\em arXiv preprint arXiv:1701.06659}, 2017.

\bibitem{girshick2015fast}
R.~Girshick.
\newblock Fast r-cnn.
\newblock In {\em {Proc. of Computer Vision and Pattern Recognition (CVPR)}},
  2015.

\bibitem{girshick2014rich}
R.~Girshick, J.~Donahue, T.~Darrell, and J.~Malik.
\newblock Rich feature hierarchies for accurate object detection and semantic
  segmentation.
\newblock In {\em {Proc. of Computer Vision and Pattern Recognition (CVPR)}},
  2014.

\bibitem{hariharan2015hypercolumns}
B.~Hariharan, P.~Arbel{\'a}ez, R.~Girshick, and J.~Malik.
\newblock Hypercolumns for object segmentation and fine-grained localization.
\newblock In {\em {Proc. of Computer Vision and Pattern Recognition (CVPR)}},
  2015.

\bibitem{he2016deep}
K.~He, X.~Zhang, S.~Ren, and J.~Sun.
\newblock Deep residual learning for image recognition.
\newblock In {\em {Proc. of Computer Vision and Pattern Recognition (CVPR)}},
  2016.

\bibitem{huang2017densely}
G.~Huang, Z.~Liu, K.~Q. Weinberger, and L.~van~der Maaten.
\newblock Densely connected convolutional networks.
\newblock In {\em {Proc. of Computer Vision and Pattern Recognition (CVPR)}},
  2017.

\bibitem{huang2017speed}
J.~Huang, V.~Rathod, C.~Sun, M.~Zhu, A.~Korattikara, A.~Fathi, I.~Fischer,
  Z.~Wojna, Y.~Song, S.~Guadarrama, et~al.
\newblock Speed/accuracy trade-offs for modern convolutional object detectors.
\newblock In {\em {Proc. of Computer Vision and Pattern Recognition (CVPR)}},
  2017.

\bibitem{jisoo2017rssd}
J.~Jeong, H.~Park, and N.~Kwak.
\newblock Enhancement of ssd by concatenating feature maps for object
  detection.
\newblock In {\em BMVC}, 2017.

\bibitem{kong2016hypernet}
T.~Kong, A.~Yao, Y.~Chen, and F.~Sun.
\newblock Hypernet: Towards accurate region proposal generation and joint
  object detection.
\newblock In {\em {Proc. of Computer Vision and Pattern Recognition (CVPR)}},
  2016.

\bibitem{krizhevsky2012imagenet}
A.~Krizhevsky, I.~Sutskever, and G.~E. Hinton.
\newblock Imagenet classification with deep convolutional neural networks.
\newblock In {\em {Proc. of Neural Information Processing Systems (NIPS)}},
  2012.

\bibitem{jianan2015scaleaware}
J.~Li, X.~Liang, S.~Shen, T.~Xu, J.~Feng, and S.~Yan.
\newblock Scale-aware fast r-cnn for pedestrian detection.
\newblock In {\em arXiv preprint arXiv:1510.08160}, 2015.

\bibitem{lin2017feature}
T.-Y. Lin, P.~Doll{\'a}r, R.~Girshick, K.~He, B.~Hariharan, and S.~Belongie.
\newblock Feature pyramid networks for object detection.
\newblock In {\em {Proc. of Computer Vision and Pattern Recognition (CVPR)}},
  2017.

\bibitem{lin2014microsoft}
T.-Y. Lin, M.~Maire, S.~Belongie, J.~Hays, P.~Perona, D.~Ramanan,
  P.~Doll{\'a}r, and C.~L. Zitnick.
\newblock Microsoft coco: Common objects in context.
\newblock In {\em {Proc. of European Conf. on Computer Vision (ECCV)}}, 2014.

\bibitem{liu2016ssd}
W.~Liu, D.~Anguelov, D.~Erhan, C.~Szegedy, S.~Reed, C.-Y. Fu, and A.~C. Berg.
\newblock Ssd: Single shot multibox detector.
\newblock In {\em {Proc. of European Conf. on Computer Vision (ECCV)}}, 2016.

\bibitem{liu2015parsenet}
W.~Liu, A.~Rabinovich, and A.~C. Berg.
\newblock Parsenet: Looking wider to see better.
\newblock {\em arXiv preprint arXiv:1506.04579}, 2015.

\bibitem{long2015fully}
J.~Long, E.~Shelhamer, and T.~Darrell.
\newblock Fully convolutional networks for semantic segmentation.
\newblock In {\em {Proc. of Computer Vision and Pattern Recognition (CVPR)}},
  2015.

\bibitem{luo2016understanding}
W.~Luo, Y.~Li, R.~Urtasun, and R.~Zemel.
\newblock Understanding the effective receptive field in deep convolutional
  neural networks.
\newblock In {\em {Proc. of Neural Information Processing Systems (NIPS)}},
  2016.

\bibitem{newell2016stacked}
A.~Newell, K.~Yang, and J.~Deng.
\newblock Stacked hourglass networks for human pose estimation.
\newblock In {\em {Proc. of European Conf. on Computer Vision (ECCV)}}, 2016.

\bibitem{noh2015learning}
H.~Noh, S.~Hong, and B.~Han.
\newblock Learning deconvolution network for semantic segmentation.
\newblock In {\em {Proc. of Computer Vision and Pattern Recognition (CVPR)}},
  2015.

\bibitem{redmon2016you}
J.~Redmon, S.~Divvala, R.~Girshick, and A.~Farhadi.
\newblock You only look once: Unified, real-time object detection.
\newblock In {\em {Proc. of Computer Vision and Pattern Recognition (CVPR)}},
  2016.

\bibitem{redmon2017yolo9000}
J.~Redmon and A.~Farhadi.
\newblock Yolo9000: better, faster, stronger.
\newblock In {\em {Proc. of Computer Vision and Pattern Recognition (CVPR)}},
  2017.

\bibitem{ren2017accurate}
J.~Ren, X.~Chen, J.~Liu, W.~Sun, J.~Pang, Q.~Yan, Y.-W. Tai, and L.~Xu.
\newblock Accurate single stage detector using recurrent rolling convolution.
\newblock In {\em {Proc. of Computer Vision and Pattern Recognition (CVPR)}},
  2017.

\bibitem{ren2015faster}
S.~Ren, K.~He, R.~Girshick, and J.~Sun.
\newblock Faster r-cnn: Towards real-time object detection with region proposal
  networks.
\newblock In {\em {Proc. of Neural Information Processing Systems (NIPS)}},
  2015.

\bibitem{sermanet2013overfeat}
P.~Sermanet, D.~Eigen, X.~Zhang, M.~Mathieu, R.~Fergus, and Y.~LeCun.
\newblock Overfeat: Integrated recognition, localization and detection using
  convolutional networks.
\newblock In {\em {Proc. of Int'l Conf. on Learning Representations (ICLR)}},
  2013.

\bibitem{simonyan2014very}
K.~Simonyan and A.~Zisserman.
\newblock Very deep convolutional networks for large-scale image recognition.
\newblock In {\em {Proc. of Int'l Conf. on Learning Representations (ICLR)}},
  2014.

\bibitem{szegedy2017inception}
C.~Szegedy, S.~Ioffe, V.~Vanhoucke, and A.~A. Alemi.
\newblock Inception-v4, inception-resnet and the impact of residual connections
  on learning.
\newblock In {\em AAAI}, 2017.

\bibitem{szegedy2015going}
C.~Szegedy, W.~Liu, Y.~Jia, P.~Sermanet, S.~Reed, D.~Anguelov, D.~Erhan,
  V.~Vanhoucke, and A.~Rabinovich.
\newblock Going deeper with convolutions.
\newblock In {\em {Proc. of Computer Vision and Pattern Recognition (CVPR)}},
  2015.

\bibitem{tulsiani2015viewpoints}
S.~Tulsiani and J.~Malik.
\newblock Viewpoints and keypoints.
\newblock In {\em {Proc. of Computer Vision and Pattern Recognition (CVPR)}},
  2015.

\bibitem{uijlings2013selective}
J.~R. Uijlings, K.~E. Van De~Sande, T.~Gevers, and A.~W. Smeulders.
\newblock Selective search for object recognition.
\newblock In {\em {Int'l Journal of Computer Vision (IJCV)}}, volume 104, pages
  154--171, 2013.

\bibitem{xiang2017context}
W.~Xiang, D.-Q. Zhang, V.~Athitsos, and H.~Yu.
\newblock Context-aware single-shot detector.
\newblock {\em arXiv preprint arXiv:1707.08682}, 2017.

\bibitem{xie2017aggregated}
S.~Xie, R.~Girshick, P.~Doll{\'a}r, Z.~Tu, and K.~He.
\newblock Aggregated residual transformations for deep neural networks.
\newblock In {\em {Proc. of Computer Vision and Pattern Recognition (CVPR)}},
  2017.

\bibitem{junjie2013mtdpm}
J.~Yan, X.~Zhang, Z.~Lei, S.~Liao, and S.~Z. Li.
\newblock Robust multi-resolution pedestrian detection in traffic scenes.
\newblock In {\em CVPR}, 2013.

\bibitem{yu2016multi}
F.~Yu and V.~Koltun.
\newblock Multi-scale context aggregation by dilated convolutions.
\newblock In {\em {Proc. of Int'l Conf. on Learning Representations (ICLR)}},
  2016.

\bibitem{zhou2015object}
B.~Zhou, A.~Khosla, A.~Lapedriza, A.~Oliva, and A.~Torralba.
\newblock Object detectors emerge in deep scene cnns.
\newblock In {\em {Proc. of Int'l Conf. on Learning Representations (ICLR)}},
  2015.

\bibitem{zhu2017couplenet}
Y.~Zhu, C.~Zhao, J.~Wang, X.~Zhao, Y.~Wu, and H.~Lu.
\newblock Couplenet: Coupling global structure with local parts for object
  detection.
\newblock In {\em {Proc. of Int'l Conf. on Computer Vision (ICCV)}}, 2017.

\end{thebibliography}
}

\end{document}